\documentclass[%preprint,
preprintnumbers,
superscriptaddress,
footinbib,
amsfonts,
amssymb,
amsmath,
intlimits,
onecolumn,
]{revtex4-2}
\usepackage[utf8]{inputenc}
\usepackage[T2A]{fontenc}
\usepackage{bm,latexsym,mathrsfs,enumerate,amsmath,amssymb}
\usepackage{upgreek}
\usepackage[table,x11names]{xcolor}
\usepackage[breaklinks=true,
unicode=true,
urlcolor = blue,
colorlinks = true,
citecolor = blue,
linkcolor = blue
]{hyperref}
\usepackage[]{graphicx}
\usepackage{wasysym}
\usepackage[cal=boondoxo,scr=rsfs]{mathalfa}
\usepackage{mathtools}
\renewcommand{\vec}[1]{\mathbf{#1}}

\usepackage{makecell,tabularx,multirow}
\usepackage{hhline}

\usepackage{chemformula}

\usepackage{titlesec}

\titleformat{\section}[display]%
{\raggedright\large\bfseries}% global formatting (number and title)
{\thesection}% label: number and its formatting
{1em}% spacing between number and title
{}% optional (content between number and title)
\titlespacing*{\section}
{0pt}% left
{1.3cm}% before
{0pt}% after

\begin{document}

\title{{Field-programmable dynamics in a soft magnetic actuator enabling true random number generation and reservoir computing}}

\author{Eduardo Sergio~Oliveros-Mata}
\thanks{These authors contributed equally}
\affiliation{Helmholtz-Zentrum Dresden-Rossendorf e.V., Institute of Ion Beam Physics and Materials Research, 01328 Dresden, Germany}

\author{Oleksandr V.~Pylypovskyi}
\thanks{These authors contributed equally}
\affiliation{Helmholtz-Zentrum Dresden-Rossendorf e.V., Institute of Ion Beam Physics and Materials Research, 01328 Dresden, Germany}
\affiliation{Kyiv Academic University, 03142 Kyiv, Ukraine}

\author{Eleonora~Raimondo}
\thanks{These authors contributed equally}
\affiliation{Department of Mathematical and Computer Sciences, Physical Sciences and Earth Sciences, University of Messina, 98166 Messina, Italy}
\affiliation{Istituto Nazionale di Geofisica e Vulcanologia, 00143 Rome, Italy}

\author{Rico~Illing}
\affiliation{Helmholtz-Zentrum Dresden-Rossendorf e.V., Institute of Ion Beam Physics and Materials Research, 01328 Dresden, Germany}

\author{Yevhen~Zabila}
\affiliation{Helmholtz-Zentrum Dresden-Rossendorf e.V., Institute of Ion Beam Physics and Materials Research, 01328 Dresden, Germany}

\author{Lin~Guo}
\affiliation{Helmholtz-Zentrum Dresden-Rossendorf e.V., Institute of Ion Beam Physics and Materials Research, 01328 Dresden, Germany}

\author{Guannan~Mu}
\affiliation{Helmholtz-Zentrum Dresden-Rossendorf e.V., Institute of Ion Beam Physics and Materials Research, 01328 Dresden, Germany}

\author{M\'{o}nica~Navarro~L\'{o}pez}
\affiliation{Helmholtz-Zentrum Dresden-Rossendorf e.V., Institute of Ion Beam Physics and Materials Research, 01328 Dresden, Germany}

\author{Xu~Wang}
\affiliation{Helmholtz-Zentrum Dresden-Rossendorf e.V., Institute of Ion Beam Physics and Materials Research, 01328 Dresden, Germany}
\affiliation{Helmholtz-Zentrum Hereon, Institute of Functional Materials for Sustainability, 14513 Teltow, Germany}
%\homepage{current address}

\author{Georgios~Tzortzinis}
\affiliation{Technische Universit\"{a}t Dresden, Dresden Center for Intelligent Materials (DCIM), 01307 Dresden, Germany}
\affiliation{Technische Universit\"{a}t Dresden, Institute of Lightweight Engineering and Polymer Technology, 01307 Dresden, Germany}

\author{Angelos~Filippatos}
\affiliation{Department of Mechanical Engineering \& Aeronautics, Machine Design Laboratory, University of Patras, 26504 Patras, Greece}

\author{Gilbert~Santiago~Ca\~{n}\'{o}n~Berm\'{u}dez}
\affiliation{Helmholtz-Zentrum Dresden-Rossendorf e.V., Institute of Ion Beam Physics and Materials Research, 01328 Dresden, Germany}

\author{Francesca~Garesc\`{\i}}
\affiliation{Department of Engineering, University of Messina, 98166 Messina, Italy}

\author{Giovanni~Finocchio}
\email{gfinocchio@unime.it}
\affiliation{Department of Mathematical and Computer Sciences, Physical Sciences and Earth Sciences, University of Messina, 8166 Messina, Italy}

\author{Denys~Makarov}
\email{d.makarov@hzdr.de}
\affiliation{Helmholtz-Zentrum Dresden-Rossendorf e.V., Institute of Ion Beam Physics and Materials Research, 01328 Dresden, Germany}

\begin{abstract} 
	Complex and even chaotic dynamics, though prevalent in many natural and engineered systems, has been largely avoided in the design of electromechanical systems due to concerns about wear and controlability. Here, we demonstrate that complex dynamics might be particularly advantageous in soft robotics, offering new functionalities beyond motion not easily achievable with traditional actuation methods. We designed and realized resilient magnetic soft actuators capable of operating in a tunable dynamic regime for tens of thousands cycles without fatigue. We experimentally demonstrated the application of these actuators for true random number generation and stochastic computing. {W}e validate soft robots as physical reservoirs capable of performing Mackey--Glass time series prediction. These findings show that exploring the complex dynamics in soft robotics would extend the application scenarios in soft computing, human-robot interaction and collaborative robots as we demonstrate with biomimetic blinking and randomized voice modulation.
\end{abstract}

\maketitle

{A large number of mechanical systems, including simple ones such as the double pendulum, exhibit dynamics characterized by deterministic periodic and chaotic responses depending on the excitation frequency $f$ and amplitude $A$ of the applied force}~\cite{marszal2014bifurcations}. Mechanical systems with a tendency to chaotisation demonstrate multiple resonances and various transitions to chaos~\cite{Touze11}. Today, the concept of complexity and, especially, deterministic chaos that refers to systems without stochastic fluctuations jet losing stability of phase space trajectories is explored for a variety of directions~\cite{Crutchfield12} even including biological systems~\cite{Faber18} or optics~\cite{Fan21}. In particular, chaos is a fundamental aspect of electromechanical systems and is broadly explored in motion planning for mobile rigid robots, fluid mixing, and improving energy harvesting, as well as in mechanisms used in washing machines, dishwashers, and air conditioners~\cite{kundu2023nonlinear}. Although the analysis of traditional robotics and mechanisms has revealed inherent chaotic dynamics \cite{zang2016applications}, chaos can also be intentionally generated through nonlinear feedback~\cite{kundu2023nonlinear} to achieve specific functionalities. % However, chaos can also exacerbate wear in systems involving gears, bearings, and other rigid components. The increased mechanical wear due to abrasion, corrosion, fatigue, and adhesion~\cite{burwell1957survey}, are amplified as operating conditions fluctuate during chaotic motion~\cite{geng2021bifurcation}.

In contrast to rigid mechanisms, soft actuators can facilitate transition into complex dynamics without the need for dedicated feedback algorithms. Mechanically soft actuators do not possess any rigid components in their embodiment rendering them ideally suited to explore complex and even chaotic dynamics which is typically observed at higher frequencies ({Supplementary Tables~1 and~2}).  
The inherent nonlinear oscillations emerging in soft actuators for specific parameter values ~\cite{cerino2016role,zhang2024soft} can be applied for secure, biomimetic, and soft computing applications. However, as soft actuators exhibit low mechanical impedance, their control at high frequencies is challenging~\cite{tang2020leveraging,Chin20}. Therefore, until now, most high-speed actuators emphasize the benefits of deterministic motion~\cite{li2017fast,hu2018small,kim2018printing,ren2019multi,duduta2019realizing,tang2020leveraging,chortos20203d,mao2022ultrafast,Qing2024}, with few incorporating resonant movement~\cite{chen2019controlled,wu2019insect,han2022ultrathin}, and even explicitly avoid nonlinear kinematic modes~\cite{chen2019controlled}. Despite the potential to enhance the capabilities of soft actuators, chaotic behavior has been avoided~\cite{dai2016effect,zhao2023vibrations}, and infrequently exploited~\cite{kumar2016chaotic,kim2020random,kim2024cryptographic} for the study of complex dynamics in soft actuators.

Avoiding mechanical constraints of traditional mechanisms, soft actuators have been developed to operate even untethered, for cases where high actuation speeds and flexibility are essential. The untethered approach allows for unobstructed motion at high actuation speeds, eliminating the mechanical limitations imposed by connectors, hoses, or cables. Untethered actuation would allow soft actuators to utilize unhindered their available degrees of freedom leveraging complex dynamics for advanced functionalities beyond motion. Among others, magnetic soft actuators allow unobstructed remote operation that can be controlled through magnetic fields~\cite{wu2020multifunctional}. Owing to these capabilities, magnetic flexible devices have been demonstrated for applications such as drug treatment, exploration, and surgery in delicate remote internal body regions~\cite{hwang2020review,oliveros2023magnetically}. Swimming, walking, rotating, jumping, and levitating become simpler when using untethered magnetic actuation~\cite{wang2020untethered,chung2021magnetically}. Additionally, they can be structurally programmed~\cite{deng2020laser}, making them compatible with smart machine concepts. However, {as true random number generation, stochastic computing, and reservoir computing have} not been demonstrated in untethered magnetic soft actuators. Soft computing, which prioritizes approximate solutions to {complex} problems, becomes especially relevant for soft robots that are not ideally suited to traditional precise control algorithms~\cite{shukla2000soft}.

Here, we experimentally validate that ultrathin magnetic soft robots can exhibit tunable complex dynamics when excited by time-varying magnetic fields. By adjusting the frequency and amplitude of the magnetic field, we can tailor the dynamics of soft actuators from periodic, through quasiperiodic, to chaotic regimes, offering a new dimension in the control of soft actuators.  We use tetherless magnetic actuation on a highly resilient magnetic composite which allows reliable actuation for over continuous tens of thousands of cycles even in a chaotic regime. Our soft actuators showcase remarkable reliability comparable to the standards of traditional robotics even operating at high speeds. {Unlike typical examples of alternating-field-driven magnetic soft actuators that show regular (periodic) or repeatable nonlinear motion, our actuators show complex dynamic behavior under a simple sinusoidal field drive. Because of self-interaction (contact/friction and a self-field similar to the drive), their motion becomes aperiodic and even chaotic, which we turn into true random number generation (passing standard NIST tests about the quality) and use for stochastic/neuromorphic computing (reservoir tasks).} We study a spectrum of complex dynamics of magnetic soft actuators to demonstrate novel applications in soft robotics beyond physical motion, including hardware random number generation, soft and reservoir computing. As a demonstrator of soft computing we realized stochastic multiplication, while the soft-robot-based physical reservoir was applied for predictive tasks using harmonic and non-harmonic inputs { and wave transformation}.

\section*{Results and Discussion}

\subsection*{Complex dynamics of magnetic soft robots} 

\begin{figure}[h]
    \centering
    \includegraphics[width=\textwidth]{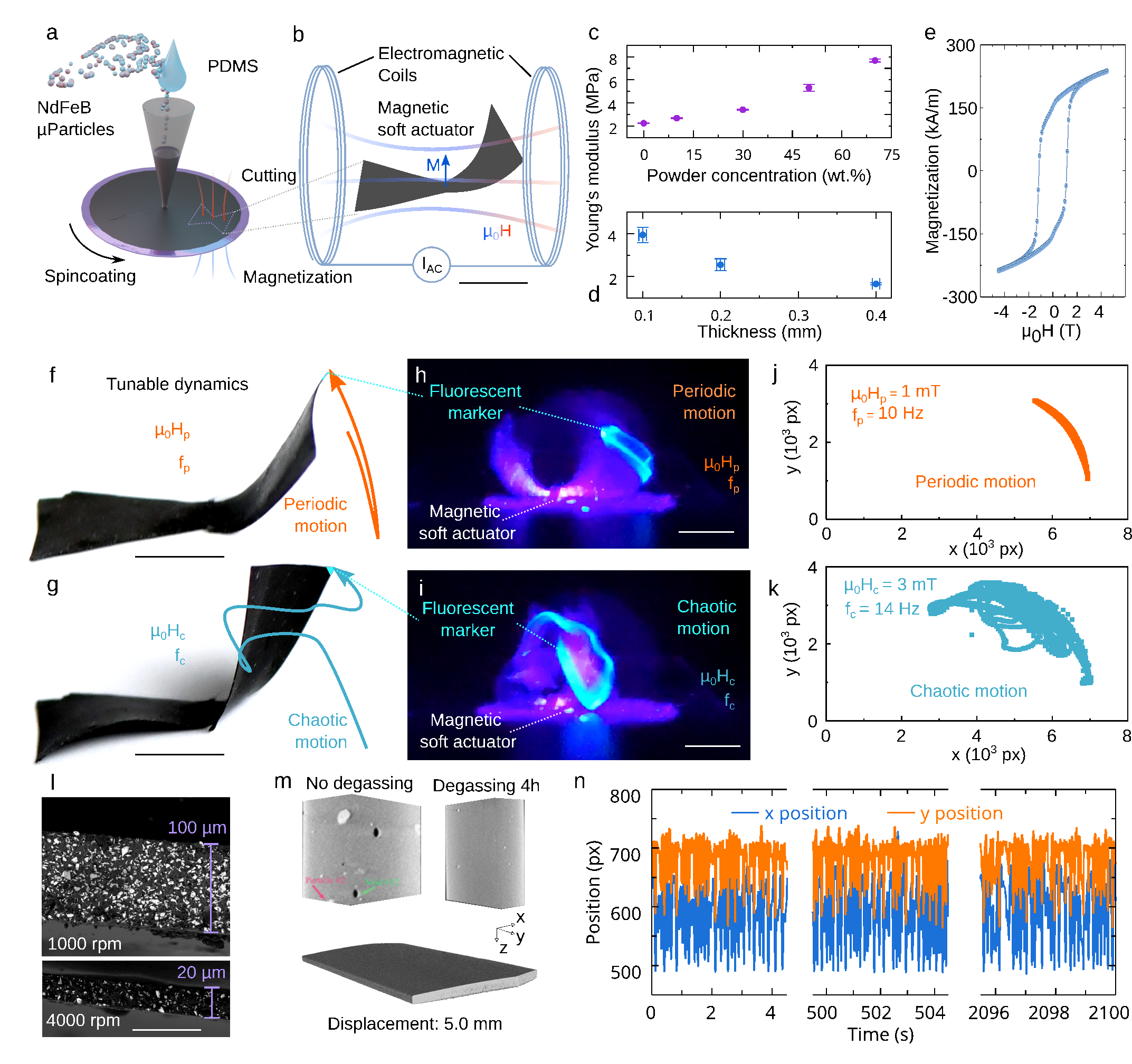} 
    \caption{\textbf{Chaotic behavior in classical systems and its translation to magnetic soft actuators.}  (a) Schematic of fabrication of magnetic actuators, ({b}) particularly tailored to actuate in alternating magnetic fields. ({c}) Young's modulus of the composite for distinct magnetic powder concentration {(400-$\mu$m-thick membranes of NdFeB in PDMS, 10:1 ratio base to curing agent)} and ({d}) sample thickness {(NdFeB in PDMS, 20:1 ratio of base to curing agent)}. ({e}) Magnetic hysteresis curve of the composite. ({f}) In the periodic dynamics region, the magnetic actuators move at the driving frequency, creating flapping-like movements. ({g}) Under certain field intensities and frequencies, the actuators show unpredictable chaotic motion. Tracking of the position of a fluorescent marker on the body of the soft actuators ({h},\,{i}) during ({j}) periodic and ({k}) chaotic dynamic modes controlled by the frequency and intensity of the applied magnetic fields. ({l}) Cross sections and ({m}) computed tomography images of the resilient composite withstanding ({n})~40\,000~cycles in the chaotic dynamics regime.}
    \label{fig:pendulum} %
\end{figure}

Figure~\ref{fig:pendulum}a illustrates the fabrication process based of spin coating of the magnetic composite based on polydimethylsiloxane (PDMS) containing 70\,wt.\% NdFeB hard magnetic nanoparticles, magnetization and cutting. The actuator consists of a magnetic composite 80-$\mu$m-thick membrane. We produced two types of robots with different shapes: rectangular and  a bow tie with a 2-mm-wide neck (Figure~\ref{fig:pendulum}b), with the lateral sizes of 15\,mm\,$\times$\,5\,mm and 13\,mm\,$\times$\,8\,mm, respectively. The magnetocrystalline anisotropy aligns the soft robot magnetization along the out-of-plane direction ($\vec{M}$ direction in Figure~\ref{fig:pendulum}b). This composite material is  tested to withstand tens of thousands continuous cycles in chaotic dynamic regimes. Higher particle concentrations enhance the magnetic response but result in an increase of the Young's modulus (Figure~\ref{fig:pendulum}c). Yet, the Young's modulus is lowered with increasing the thickness of the membrane (Figure~\ref{fig:pendulum}d). Using NdFeB particles allowed us to obtain a hard magnetic composite that enabled an efficient actuation using magnetic fields in the mT range without unwanted remagnetization processes  (Figure~\ref{fig:pendulum}e). 

We have characterized the dynamical properties of soft robots actuated with AC magnetic field generated by electromagnetic coils supplied by an AC current oriented along the lying body direction as shown in Figure~\ref{fig:pendulum}b. The resilience of {magnetic membranes of different thickness (from 20 to 100\,$\mu$m)} was achieved by minimizing structural inhomogeneities {e.g.} voids or cracks (Figure~\ref{fig:pendulum}l). As confirmed by computed tomography (CT) scans (Figures~\ref{fig:pendulum}m,~Supplementary Figures~1,2) a degassing process at 0.1\,mbar for 4\,h reduces the number of voids of the films (Supplementary Figure~3, Supplementary Table~3).  These soft actuators withstand more than 40\,000 of actuation cycles in following patters as of the chaotic dynamics due to excitation of large number of vibrational modes~\cite{Touze11}  (Figure~\ref{fig:pendulum}n). {It is remarkable that even driven in the highly nonlinear (chaotic) regime for thousands of cycles, the actuator performs well without any noticeable fatigue. Although this mechanical stability is well established for regular periodic motion of magnetic soft actuators, our demonstration enables use cases where these complex and robust dynamics is used for computing.}

The different dynamic modes were studied by applying alternating magnetic fields at selected frequencies (from 1\,Hz to 20\,Hz) and intensities (from 0.5\,mT to 9\,mT). This generates a lifting motion of the arms of the actuator that follow trajectories of a different level of nonlinearity from periodic response (Figure~\ref{fig:pendulum}f) to non-periodic response (Figure~\ref{fig:pendulum}g). We applied fluorescent markers at the apex of the arms of the magnetic soft actuators (Figure~\ref{fig:pendulum}h,i) to enable real time tracking of the position of robotic arms and allow to identify the motion regimes through the analysis of the position data. The frame-by-frame position of the marker allowed to visualize trajectories of the actuator during these fast processes. As an example, we show the up and down movement of an actuator in a periodic motion regime subjected to a sinusoidal magnetic field of 1\,mT at 10\,Hz for 2\,s (Figure~\ref{fig:pendulum}j). In this case, trajectories of the tracked point repeat at each actuation period. On the other hand, complex unpredictable dynamics appear after changing the parameters of the field to 3\,mT at 14\,Hz (Figure~\ref{fig:pendulum}k) because of the excitation of additional vibrational modes. Below we show that this motion corresponds to chaotic dynamics. By slightly changing the parameters of the sinusoidal input field, we can obtain different dynamics without complex feedback algorithms, that we explore in the following sections providing an analysis of the movement and addressing potential application scenarios.

\begin{figure}
    \centering
    \includegraphics[width=\textwidth]{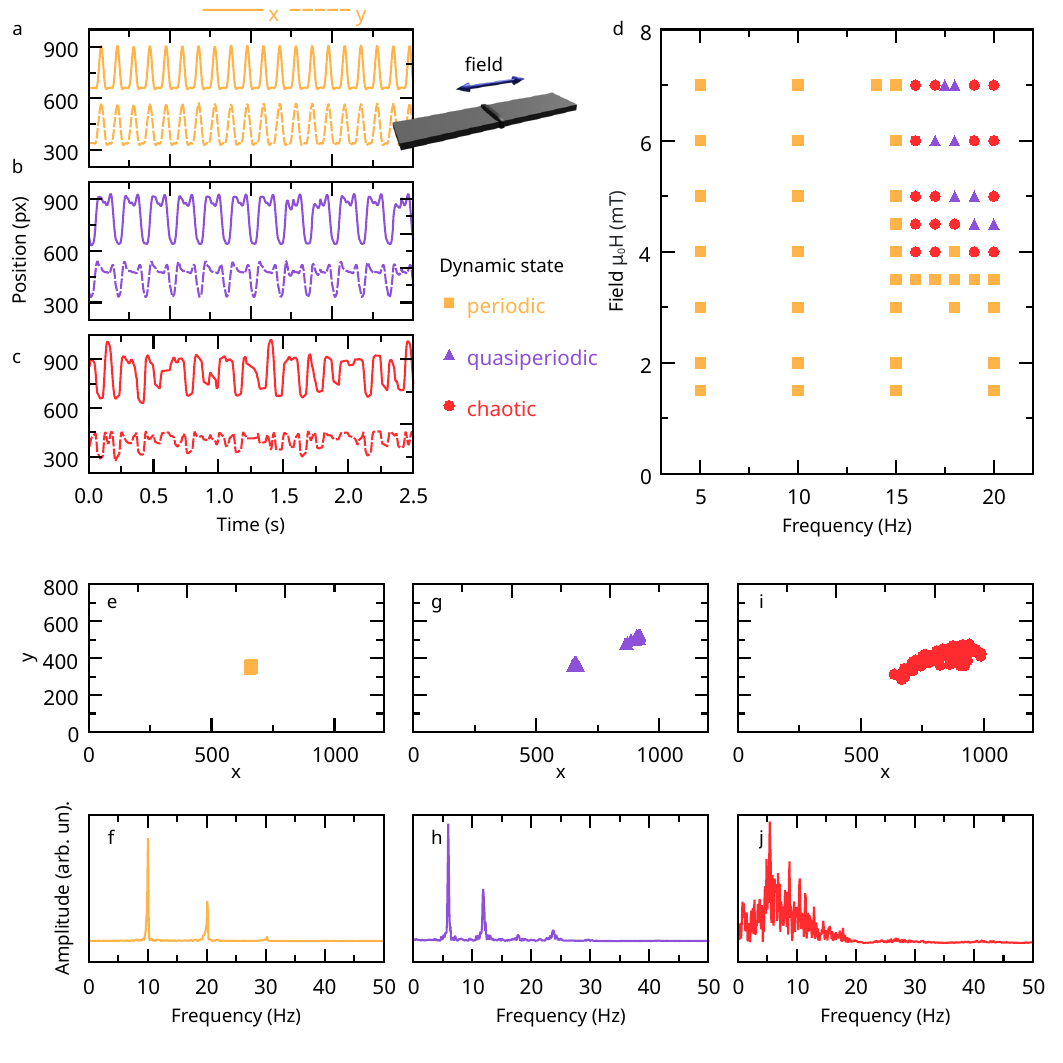}
    \caption{\textbf{Dynamic states of a rectangular-shape magnetic soft actuator in alternating magnetic fields along its long axis.} Tracking magnetic soft actuator motion over time in alternating magnetic fields of specific intensity and frequency, demonstrating ({a}) periodic {(10\,Hz, 5\,mT)}, ({b}) quasiperiodic {(18\,Hz, 5\,mT)}, and ({c}) chaotic behavior {(16\,Hz, 5\,mT) of the actuator schematically shown as an inset in panel (a)}. For the latter, no clear pattern is observed. Solid ({dashed}) lines show the time evolution of the $x$ ($y$) coordinates of the arm of the soft robot. ({d}) Dynamic states diagram across varying frequencies and intensities of the driving magnetic field. Poincar\'{e} diagrams and Fourier spectra of the actuator's $x$-position for ({e,\,f}) periodic, ({g,\,h}) quasiperiodic, and ({i,\,j}) chaotic behavior corresponding to panels ({a}), ({b}), and ({c}), respectively. }
    \label{fig:dynamic_states} 
\end{figure}

We performed a comprehensive analysis of the effect of the field intensity and frequency on the dynamics of the magnetic soft actuators. Figure~\ref{fig:dynamic_states} shows the position of the tracked fluorescent point on the soft rectangular actuator during actuation along it's long axis and designation of the dynamic regimes for specific parameters (see also {Supplementary Figure~4, Supplementary Movies S1 and S2}). In a classical flexible magnetic lifter, deformations are typically proportional to the applied field intensity. For this reason, an oscillating magnetic field is expected to generate a synchronized deflection of the actuator. This behavior, referred to as the periodic regime, is shown in Figure~\ref{fig:dynamic_states}a {(10\,Hz, 5\,mT)}, where the position of the fluorescent tracker moves harmonically in response to the applied magnetic field. An increase of power delivered by the actuation {via field intensity or frequency} enables resonances on subharmonics making the dynamic profile more complex but still predictable, which is referred to as the quasiperiodic regime (Figure~\ref{fig:dynamic_states}b{, 18\,Hz, 10\,mT}). In this regime, the actuator exhibits a cross-clapping-like motion, alternating the position of its upper arm~\cite{wang2020untethered}. Finally, chaotic dynamics were observed, characterized by irregular motion patterns that could not be predicted from the previous cycles of the alternating magnetic field (Figure~\ref{fig:dynamic_states}c{, 16\,Hz, 5\,mT}). This is a result of entering into a strongly nonlinear motion regime related to the excitation of high order and strongly non-uniform mechanical modes. At higher field intensities, the deflections become large, leading the actuator to interact with its own stray fields, which are of the same order of magnitude as the input fields. These self-interactions, added to a wide range of possible mechanical configurations of the soft actuator, enforce the complexity of chaotic dynamics which is those system is then due to a nonlinear magneto-mechanical coupling. As shown in Figure~\ref{fig:dynamic_states}d, periodic dynamics dominate at smaller field intensities and frequencies (linear response), but once the field exceeds {4}\,mT and {15}\,Hz, nonlinearities in the systems drive non-periodic dynamics. In  this higher frequency and intensity region, the balance between all the possible actuator configurations and the strength of the interactions is more subtle, leading to regions of chaotic dynamics.

The classification of the motion regimes was done using Poincar\'{e} diagrams for tracked points and spectral analysis of the arm position vs time. In particular, Poincar\'{e} diagrams represent sampling of a trajectory of the actuator’s end with the frequency of the actuating field. In a periodic regime, a Poincar\'{e} diagram reveals several well-defined clusters of points independent of the number of tracked actuation periods (Figure~\ref{fig:dynamic_states}e). The actuator’s end always visits a static set of points on the Poincar\'{e} diagram over multiple actuating periods, while the spatial configuration is dependent only on the initial sampling time. The respective Fourier spectrum of the coordinate shows equidistant peaks at the excitation frequency  (Figure~\ref{fig:dynamic_states}f) confirming that this motion is periodic. A quasiperiodic motion regime is characterized by the appearance of more clusters in a Poincar\'{e} diagram (Figure~\ref{fig:dynamic_states}g) and a broadening of the corresponding Fourier spectrum by resonances on subharmonics and their multiples still preserving discreteness of the spectrum (Figure~\ref{fig:dynamic_states}h). The chaotic motion can be clearly distinguished from the above discussed regions relying on the analysis of the Poincar\'{e} diagram, which does not contain clustering in favor of drawing of a strange attractor (Figure~\ref{fig:dynamic_states}i). An instability of the phase of the trajectory (Figure~\ref{fig:dynamic_states}c) leads to the dense covering of a certain region in the Poincar\'{e} diagram by points which never coincide indicating unpredictability of the motion pattern during each input cycle and divergence of the trajectories in the phase space. The respective Fourier spectrum shows a continuous range of excited harmonics (Figure~\ref{fig:dynamic_states}j). This analysis allowed us to confirm the presence of different dynamic modes, and, for the first time, the presence of chaotic behavior in magnetic soft actuators. {We performed the same analysis of the dynamics for a bow-tie-shaped magnetic soft actuator (Supplementary Figure~5) and and rectangular actuator with the magnetic field tilted by $30^\circ$ from the actuator's axis (Supplementary Figure~6). The difference between these datasets is quantitative related to the different amount of magnetic material and magnetic field strength delivered to the robot, but not qualitative: all robots show similar signatures of dynamic regimes and placement of the dynamic regions in Frequency-Field coordinates on their phase diagrams.}

Long-term dynamics of the soft robot provides detailed structure of the attractor at the Poincar\'{e} diagram (Supplementary Figure~7 with 41\,000 points shown). There are several elongated areas with the denses placement of points with the background of randomly scattered single points that came due to noise. Comparison of the starting and final parts of the dataset used for Supplementary Figure~7a (Supplementary Figure~7b,c) shows that the main features of the attractor are stable in time.

Quantitatively the phase diagram depends on the soft robot's shape and applied field direction. Still, the presence of three distinct dynamic regimes is robust with respect to the system parameters. Compared to the rectangular-shape robots, the phase diagram for the bow-tie-shape robots is shifted to smaller magnetic fields (c.f. Figure~\ref{fig:dynamic_states} and Supplementary Figure~5). At the same time, a tilt of the magnetic field by $30^\circ$ in plane from the axis of the robot slightly increases the critical fields to enter the chaotic regime (Supplementary Figure~6). Furthermore, for this tilted field configuration we observe a change in the pattern of chaotic dynamics. For the field aligned with the robot's axis, we observe a dense Poincar\'{e} map (c.f. Figure~\ref{fig:dynamic_states}i and Supplementary Figure~5i) with a broad Fourier spectrum (c.f. Fig.~\ref{fig:dynamic_states}j and Supplementary Figure~5j). For the field tilted by $30^\circ$, although a broad Poincar\'{e} map is preserved (Supplementary Figure~6i), the Fourier spectrum acquires several peaks (Supplementary Figure~6j). A detailed analysis (Supplementary Figure~8) shows that this is a general feature of this configuration. It is given by the different pattern of motion of the robot. Indeed, the chaotic motion in the parallel field configuration has no temporal periodicity. In contrast when the field is tilted, the short-term periodicity of the robot's motion is preserved (c.f. Figure~\ref{fig:dynamic_states}c and Supplementary Figures~8a,d,g,j). However, the phase of the motion between oscillations frequently changes randomly leading to a dense Poincar\'{e} map with a broad noise in Fourier spectra.

\subsection*{Exploring chaotic dymamics: Stochastic computing}

{The coupling between two physics (elasticity and magnetism -- i.e. the magnetic force drives non-autonomous and nonlinear elastic response) enables a complex response of the magnetic soft actuator to the external magnetic field excitation in the dynamical regime. In particular, to qualify magnetic soft actuators for true random number generation and stochastic computing, we rely on nonlinear mechanics a system with multiple (technically infinite for soft robots) degrees of freedom, intrinsic and extrinsic random processes due to magnetic thermal noise that perturbs the magnetic texture.}

\begin{figure}
    \centering
    \includegraphics[width=\textwidth]{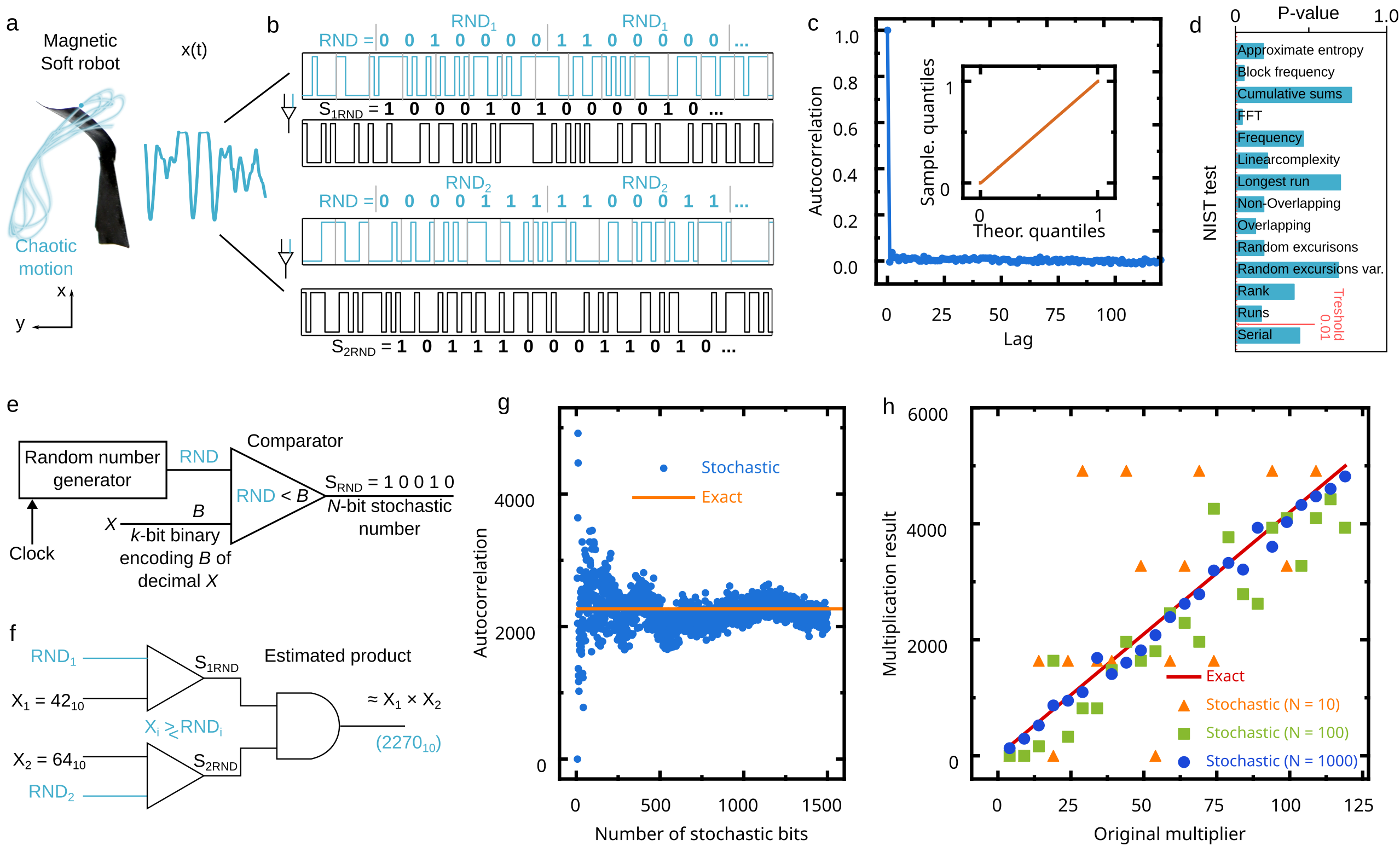} 
    \caption{\textbf{Chaotic dynamics of soft robots for stochastic soft computing.} ({a})~Time-tracked motion of a magnetic soft robot driven into chaotic dynamics. ({b})~Generation of random number sequences using the tracked x-coordinate of the soft robot. Each number is encoded in 7-bit sequences (separated by gray bars) and compared with the multiplicand to create a true random number sequence. ({c})~Rapidly decaying autocorrelation of the random number sequence used for calculations where lag refers to the distance between numbers in series. The linearity of Q-Q plot in the inset confirms that the generated random number sequence corresponds to a uniform distribution in the range $[0,1]$. ({d})~The soft-robot-generated sequence passed 14 NIST random tests (P-value threshold $= 0.01$). ({e})~Schematic of conversion of the given decimal number $B$ into $N$-bit stochastic random number. $X$ is converted to its binary representation $B$ that is compared with the sequence of $N$ random numbers RND. The output sequence is filled with 1 if $\text{RND} < B$, otherwise with 0.  ({f})~Schematic of the stochastic computing algorithm using soft-robot-generated random numbers as input. At each clock cycle, random numbers $\mathrm{RND}_1$, $\mathrm{RND}_2$ and exact numbers $X_1$, $X_2$ are compared. If $\mathrm{RND}_1 < X_1$ or $\mathrm{RND}_2 < X_2$, a stochastic bit ``1'' is generated. Otherwise, if $\mathrm{RND}_1 \ge X_1$ or $\mathrm{RND}_2 \ge X_2$, a stochastic bit ``0'' is generated.   ({g})~Stochastic multiplication of $42 \times 54$, demonstrating convergence towards the exact result (dashed line) with increasing the random bit count. ({h})~Stochastic multiplication results for different multipliers of 42 using varying stochastic bit counts ($N = 10, 100, 1000$). The red line indicates the exact result.}
    \label{fig:stochastic_computing}
\end{figure}

We demonstrate the use of magnetic soft actuators exhibiting chaotic dynamics for generating random number sequences for soft computing applications. In contrast to conventional computing, where numbers are encoded in deterministic ordered bitstreams of 0s and 1s, stochastic computing encodes numerical values as random bitstreams, where the order of the bits is irrelevant. The numerical value is encoded in the probability of observing a 1 at any given bit position. This approach enables basic arithmetic operations~-- such as multiplication~-- to be performed using simple logic elements, like AND gates, resulting in a low-cost and energy-efficient computation~\cite{Alaghi13,Alaghi18}. Unlike pseudorandom generators based on deterministic algorithms, true random number generators (TRNGs) exploit hard-to-predict variables in physical systems such as resistance, tapping time, quantum states, and noise vibration. Some flexible mechanical systems have leveraged wind-induced random motion for electricity generation and data ciphering~\cite{kim2020random,kim2024cryptographic}. Figure~\ref{fig:stochastic_computing} illustrates a soft actuator's implementation of stochastic computing using motion-generated random bits. Driven into chaotic dynamics, a magnetic soft robot creates highly unpredictable long-term patterns, producing a random data stream (Figure~\ref{fig:stochastic_computing}a). 

The normalized $x$-position of the fluorescent tracker can be transformed into a random bit stream (Figure~\ref{fig:stochastic_computing}b). Using a stochastic computing algorithm, this random bit generator estimates the product of two numbers. For the initial sequence of numbers we calculate the cumulative distribution function and use it to make the sequence uniformly distributed. To validate the randomness of the generated sequences, we analyzed the autocorrelation and distribution of the generated sequences, comparing them to a uniform random distribution. Figure~\ref{fig:stochastic_computing}c shows that the correlation between sequential numbers {close to zero} at the next number in the sequence. The smooth and linear behaviour of Q-Q plot in the inset of Figure~\ref{fig:stochastic_computing}c means that the distribution of the random numbers is well-populated in whole range and is equivalent to the uniform distribution as expected. Therefore, the obtained dataset is valid for further stochastic computing and other applications. The sequences successfully passed 14~randomness tests (Figure~\ref{fig:stochastic_computing}d) from National Institute of Science and Technology (NIST)~\cite{bassham2010statistical}. 

The random numbers are used to estimate the product of two numbers using the stochastic computing. In particular, they are utilized in the stochastic number generator (Figure~\ref{fig:stochastic_computing}e), which converts a deterministic $k$-bit binary number $B$, representing a decimal value $X$, into a stochastic bitstream $S_\text{RND}$, where the probability of a bit being $1$ approximates: $p_X \approx X/2^k$. This is achieved over $N$ clock cycles. At each cycle, a random $k$-bit number RND is generated and compared with the fixed value $B$. If $\text{RND} < B$, the outputs is 1, otherwise, it is 0. To perform stochastic multiplication, two independent $N$-bit bitstreams, $S_\text{1RND}$ and $S_\text{2RND}$, representing probabilities $p_{X_1}$ and $p_{X_2}$, are input to an AND gate (Figure~\ref{fig:stochastic_computing}f). The gate outputs a 1 only when both inputs are 1, resulting in a output stochastic bitstream with probability: $p_Z = p_{X_1} \times p_{X_2}$.

As an example, we considered two decimal values, $X_1 = 42$ and $X_2 = 54$, encoded as 7-bit binary numbers. Their respective probabilities are $p_{X_1} = 42/2^7 \approx 0.328$ and $p_{X_2} = 54/2^7 \approx 0.422$. The resulting output probability $p_Z \approx 0.138$. By scaling $p_Z$ by $2^{2k} = 2^{14}$, the resulting product is approximately 2261 that corresponds to the accuracy of 0.3\%. In Figure~\ref{fig:stochastic_computing}g we show the estimation of the product $42\times 54$ as a function of the stochastic bitstream length. As expected for stochastic algorithms, longer sequences provided better statistical accuracy and closer approximations~\cite{Wang24b}, achieving an error below 10\% for longer sequences. Figure~\ref{fig:stochastic_computing}h presents the results of stochastic multiplication between 42 and various other numbers using stochastic bitstream length $N = 10$, 100 and 1000. As the bitstream length increases, the accuracy improves: the least squares distance between the stochastic and exact results is 130 for $N = 100$, and decreases to 35 for $N=1000$.

\subsection*{Exploring quasiperiodic dynamics: Computing with a physical reservoir}

\begin{figure}[h]
    \centering
    \includegraphics[width=\textwidth]{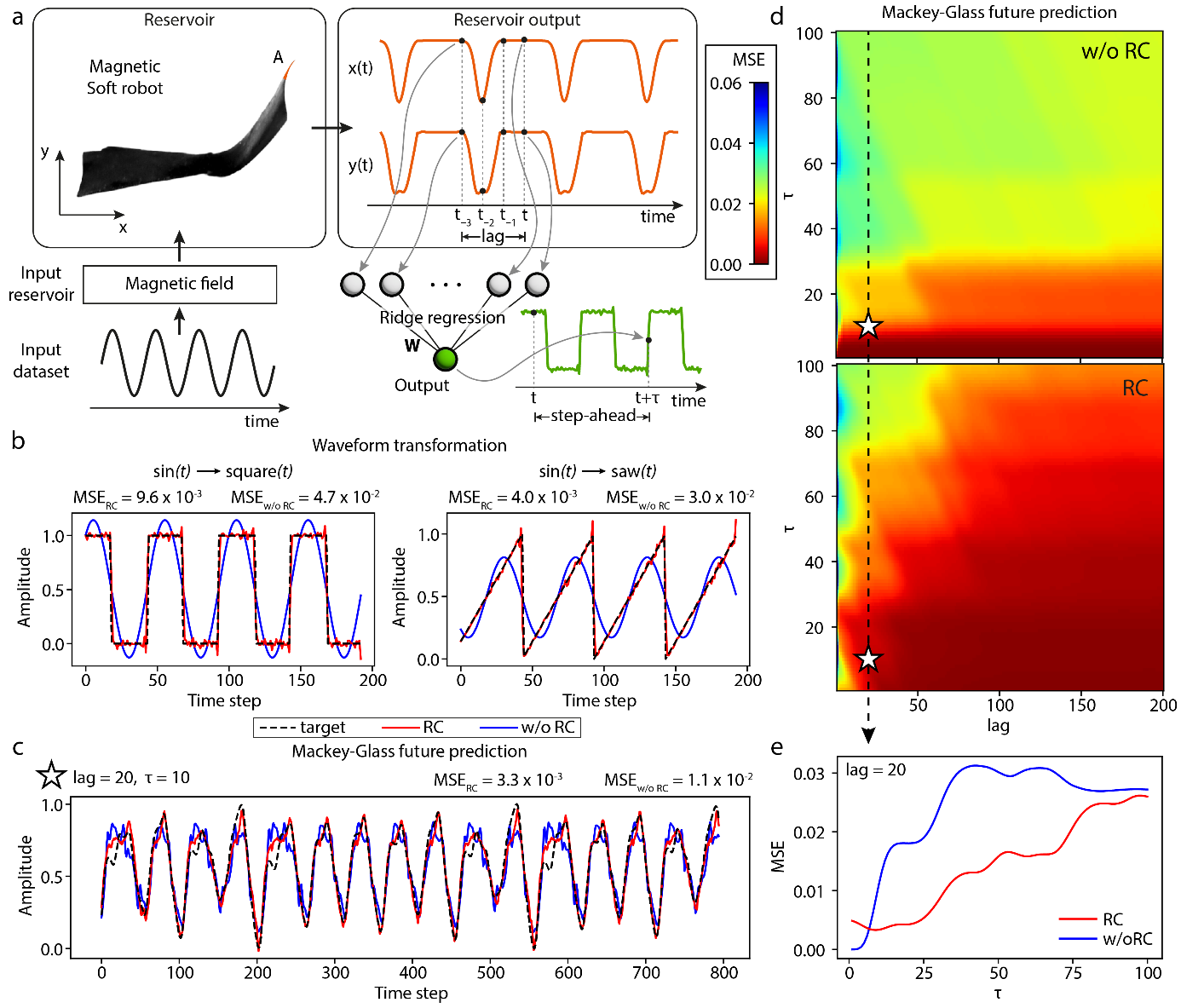}
    \caption{\textbf{Magnetic soft robot for reservoir computing (RC).} ({a}) Schematic representation of RC: an input dataset, a time-series data such as a sinusoidal function, is applied to the RC as magnetic field to drive the dynamics of the magnetic soft robot. The RC output consists of the $x(t)$ and $y(t)$ trajectories of a point at the apex of one of the robot’s arms as indicated by the orange point A. At each time step $t$, a set of past observations (lag) is used to transform or predict the time series $\tau$ step-ahead via ridge regression with trained weights {W}. ({b}) Examples of waveform transformation using the RC: a sine wave dataset, sin$(t)$, is transformed into a square wave, square$(t)$, (left) or a sawtooth wave, saw$(t)$, (right). The mean squared error (MSE) obtained with and without RC in performing this task is also reported and clearly shows the better performance of the RC. ({c}) Example of future prediction of Mackey--Glass (MG) time-series at \mbox{$\tau = 10$} and a \mbox{lag = 20}. The black dashed line is the target signal, red solid line is the prediction obtained with RC, while the blue solid line is the prediction without RC. The MSE is reported for both cases. ({d}) Complete phase diagram of the MSE (red low, blue high) of the prediction performance for MG time-series as a function of the lag and step-ahead, predictions without RC (top) and with RC (bottom) are shown. The star highlights the prediction illustrated in panel {c}. ({e}) Example of MSE as a function of step-ahead predictions for \mbox{lag = 20}, RC predictions (red line) exhibit lower MSE than those without RC (blue line) above \mbox{$\tau= 9$}.}
    \label{fig:reservoir_computing_details}
\end{figure}

We explore the use of magnetic soft robots for computing tasks focusing on reservoir computing~(RC)~\cite{Fernando03,hauser2011towards,Nakajima15,Nakajima21}. RC is a computing paradigm that received increasing attention for two main reasons: {it allows the models learn faster and with fewer data in comparison with the fully trainable deep neural network models~\cite{cisneros2022benchmarking}} and offers hardware-friendly implementations~\cite{TANAKA2019100}. In contrast to conventional deep learning systems, where computational cost scales with network depth and size, RC leverages the natural dynamics of physical systems, enabling simpler and faster training by keeping the internal structure fixed. This makes it highly suitable for real-time and embedded computing applications where computational resources are limited. The core idea of RC is to exploit complex dynamics of a nonlinear system to map input signals into a high-dimensional space where they can be processed using simple techniques such as ridge regression. A key requirement for an effective reservoir is the balance between nonlinearity and memory~\cite{Verstraeten10,Inubushi17}. The system must be sufficiently nonlinear to differentiate between similar inputs, but also possess short-term memory to preserve temporal information. In physical reservoirs, this balance is inherently governed by the material properties and external stimuli. Various physical implementations of RC have been proposed, including spintronic oscillators~\cite{Torrejon2017}, magnetic skyrmions~\cite{Sun2023} and memristor devices~\cite{Milano2022}. Here, we demonstrate a proof of concept of using a soft robot as a physical reservoir, showing its feasibility in tasks such as nonlinear waveform transformation and time-series prediction. Figure~\ref{fig:reservoir_computing_details}a shows a conceptual scheme of the proposed physical RC system. The input signals, in this case time-series data, are linearly mapped to a magnetic field that drives the dynamics of the magnetic soft robot. The nonlinear dynamic responses of the robot are captured via frame-by-frame analysis of a point at the apex of one of the robot’s arms (orange in Figure~\ref{fig:reservoir_computing_details}a), recording the $x(t)$ and $y(t)$ trajectories. These trajectories serve as the reservoir’s outputs. {Considering that we are sampling from a single point of the reservoir we have used at} each time step $t$, a set of past observations, called lag {(time-delay embeddings at the readout~\cite{Duan2023})} to transform or predict the time-series $\tau$-steps ahead, through ridge regression, which learns weights \textbf{W} to minimize the error~\cite{Parlitz2024,Duan2023,Nakajima15}. {From the hardware point of view, adding lag in a RC is less expensive than reading the configuration in different points of the soft actuator.}

We first evaluated the system's ability to perform nonlinear waveform transformations of periodic input signals~\cite{Gartside2022, Vidamour2023, Lee2024}. Specifically, we used a sine wave $\sin(t)$ as the input and transformed it into either a square wave $\mathrm{square}(t)$ or a sawtooth wave $\mathrm{saw}(t)$, as shown in Figure~\ref{fig:reservoir_computing_details}b (left and right panels, respectively). The magnetic field applied to a bow-tie-shape robot had a frequency of 4.8~Hz and an maximum intensity of 1\,mT. The performance is measured by the mean squared error (MSE) between the target signal and the output signals {(1-step-ahead prediction with a lag of 35).} The RC system successfully transformed a sine wave into the decided target waveforms with low MSE values (\mbox{MSE\textsubscript{RC}$ = 9.6 \times 10^{-3}$} and \mbox{$4.0 \times 10^{-3}$} for square and sawtooth wave, respectively). Without RC, the transformation failed and the output is simply the input signal scaled by a different amplitude, resulting in significantly higher MSE values (\mbox{MSE\textsubscript{w/oRC}$ = 4.7 \times 10^{-2}$} and \mbox{$3.0 \times 10^{-2}$}, respectively). A broader exploration of magnetic field intensities is provided in Supplementary Figure~9. In this experiments, we observed a degradation of the performance once the intensity of the field led to a chaotic robot’s response.  This demonstration highlights the ability of the soft-robot-based RC to extract nonlinear representations of input signals, enabling transformations through simple ridge regression where non-RC systems fail.

Next, we tested the magnetic soft robot RC for chaotic time-series prediction using the Mackey--Glass (MG) equations – a typical benchmark for reservoir computing~\cite{Lee2024, Yan2024, Li2024}. Figure~\ref{fig:reservoir_computing_details}c shows the prediction of MG of 10 step-ahead using input data with the lag of 20. The RC system achieved much higher accuracy than a non-RC system, \mbox{$3.3 \times 10^{-3}$} and \mbox{$1.1 \times 10^{-2}$}, respectively. A comprehensive phase diagram (Figure~\ref{fig:reservoir_computing_details}d) shows the MSE as a function of the lag and step-ahead predicted. Without RC (top panel), the performance degrades rapidly for \mbox{$\tau > 9$}, demonstrating a poor long-term prediction capability. With RC (bottom panel), MSE remains significantly lower across a broad range of lag and $\tau$ values. As an example, Figure~\ref{fig:reservoir_computing_details}e compares MSE as a function of the step-ahead prediction for a fixed lag of 20, showing that RC consistently outperformed the non-RC system for prediction up to 9 step-ahead. These results are obtained using a maximum applied magnetic field intensity of 1\,mT. A more extensive analysis of the performance across a range of magnetic field intensities is provided in Supplementary Figure~10. Additionally, we evaluated the performance of our physical RC in the frequency domain, where it once again outperformed the non-RC system. These results are provided in Supplementary Figure~11.

\section*{Discussion}

We demonstrate that the complex behavior of magnetic soft actuators offers new functionalities beyond motion replication. The excitation of chaotic dynamics driven by oscillating magnetic fields, provides a novel approach to true random number generation (TRNG). While the TRNG speed of soft robots might not be as high as some digital counterparts, it is well-suited for applications where high-speed data transmission is not paramount. Many biometric-based authentication systems operate under similarly slow bitrates, for example concepts utilize random number generation based on electrocardiographic (ECG) other biophysical signals for secure identification, even at sub-1 bit-per-second rates (Supplementary Table~4). This randomness can operate concurrently with other tasks without disrupting the data transmission or overwhelming system resources. A relatively slow generation of random numbers is particularly suitable for protocols requiring periodic re-keying, offering a secure mechanism for authentication systems without taxing device memory. This makes them ideal for scenarios where security and low computational overhead are prioritized, such as in collaborative robotic swarms or secure communication in resource-constrained environments. Additionally, the random number generation can be increased using multiple robots and swarms~\cite{wang2024heterogeneous,dong2020controlling}. Concepts involving modular robotics \cite{dong2022untethered} can expand their functionalities to include enhanced safety, for which proper system identification and triggering becomes particularly important for critical medical applications~\cite{chen2024magnetic}.

The control of the dynamical behaviour of these systems enables customization for diverse applications, from cryptography to RC. {A direction to improve the RC perfomance can be based on the use of lag with the readout of more states to have more general paradigm of reservoir computing~\cite{Gauthier21,Tavakoli24,Wang25e,Picco25}.}  In {addition}, the soft robots developed here can go beyond the simple idea of RC to work as self-adaptive RC with closed-loop architecture where the ac field driven the dynamics is given by the two contributions, one directly arising from the input and one from the current state. This latter idea can be used to develop the concept of self-learning RC~\cite{Shougat24} which identify the specific feedback to learn how to adapt the amplitude of the total external field applied to the soft robots. This configuration can enable the extraction of information coded in different time scale~\cite{Chen24b}. This approach is scalable and a number of different soft robot can be used for analyzing data having characterized by multiple time series. {Untethered dynamics in stand-alone robotics can be used in confined/biological spaces~\cite{Popov25}, where the same body used for locomotion co-integrate computation (e.g., temporal filtering or predictive analysis) without extra hardware on board. Furthermore, the same actuator at the edge that interacts with the environment supplies nonlinear dynamics and memory for reservoir computing and/or TRNG as for robot swarms with swarm signal temporal logic~\cite{Yan22a}. }

{We have demonstrated that low magnetic fields (sub-mT–few mT) and low frequencies (about $1-20$\,Hz) are necessary to control the dynamical regimes (periodic, quasi-periodic, and chaotic) of soft actuators. We note that those driving magnetic fields can be easily achieved using planar coils~\cite{Richter23}. Hence, bulky electromagnets are not necessarily needed to drive our magnetic soft robots. Furthermore, a single external field source can drive multiple magnetic soft actuators so the cost and power consumption of the external source should be mediated over the number of soft actuators. }

{Beyond computing tasks, there are further examples where the complex dynamics validated in this manuscript can find applications.} The adaptability of magnetic soft robots is crucial for mechanisms that may need to adjust their randomness characteristics based on the environment. 
Prospective soft robotics is expected to incorporate mechanisms that respond to environmental complexity to achieve full biomimicry of living-being behaviors~\cite{rus2015design}. These robots are expected to make decisions that account for safety, social compatibility, and response times that are comparable or faster than typical human reaction times. The potential of soft actuators to mimic the biological behaviour at relevant time scales is demonstrated with approximating a volunteer's eye blinking rate {(Supplementary Figure~12)} and digital voice modulation {(Supplementary Figure~13, Supplementary Movie S3)}. {Those examples show generality of the concept proposed in the manuscript.}

We expect that these examples will inspire further possible applications that integrate non-deterministic dynamics, enabling efficient exploration of soft robotic systems through complex terrains, facilitate the improvisation of new motion patterns, escape from threats using elusive maneuvers, and even for socializing including intercommunication. In this respect, soft robots can mimic rhythms and timings of biological functions, creating more natural and adaptive interactions and enhancing their social acceptance.

\section*{Methods}

\subsection{Soft actuators}
Magnetically actuated flexible membranes were fabricated by spin coating a solution of NdFeB microparticles in PDMS. NdFeB powder (MQFP-14-12-20000, Magnequench Corp.) and PDMS (Sylgard 184, Dow Corning Corp.) were mixed uniformly for 15\,min with a varied concentration of powder (0--70\,wt.\%). Curing agent (Dow Corning corp.) at {5--20} wt.\% with respect to PDMS {base} was added to the mixture and mixed for additional 10\,min. A homogeneous solution was obtained after degassing the mixture for 240\,min with a vacuum pump {to avoid bubble formation in the PDMS membrane}. We employed glass substrates with a thin coating of polyvinylalcohol (PVA, 115000, VWR) that enabled facile peeling after dissolving it in deionized water. The PVA layers were spin coated at 1000\,rpm for 30\,s and baked for 1\,h at 70\,$^\circ$C. After that, we employed a spin coating to create thin membranes of the magnetic composite. A typical speed for spin coating was between 500 to 3000\,rpm for 60\,s {to vary thickness from 50\,$\mu$m to 400\,$\mu$m (Supplementary Table 3)}. The spin coated polymers were baked at 70\,$^\circ$C for 2\,h afterwards. The obtained membranes were magnetized for 30\,s using 2.3\,T field applied in the out of plane direction. The final geometry of a soft actuator was defined by cutting using a plotter cutter (Silhouette Portrait, Silhouette America) (a bow tie shape with 13\,mm\,$\times$\,8\,mm lateral dimensions with a central 2-mm-wide neck, and a rectangle with 15\,mm\,$\times$\,5\,mm). After magnetizing and cutting the membranes, we obtained a functional actuator that can be manipulated by magnetic fields relying on the magnetic torques and forces between the magnetized composite and applied magnetic field. {Mechanical properties of the magnetic membranes cut in rectangular shape were characterized using a tensile stretching device (Supplementary Figure~15). In the main text, we focused on the 100-$\mu$m-thick sample with  70 wt.\% of NdFeB powder in PDMS (10:1 ratio of base to curing agent). The repetitive measurement of stress-strain curves for this sample is shown in Supplementary Figure~16. The consecutive curves measured to 12\% strain are fully reversible and reveal only a minor hysteresis. These measurements were carried out using tensile stretching device from SAUTER (Strain gauge: FH10).} A magnetization hysteresis loop of the {this} sample was measured through superconducting quantum interference device vibrating sample magnetometry (SQUID-VSM, Quantum Design Corp.).

\subsection{Computed tomography scans} 
Dog-bone shaped specimens are initially die-cut from circular membranes with a diameter of 50\,mm according to ISO 527-2 standard (Type 5A) (Supplementary Figure~1). The nominal gauge and overall width of the dog-bone equal to 4\,mm and 12.5\,mm, respectively, while the thickness depends on the spin-coating speed during the manufacturing stage. To address that the overall length of the extracted membrane falls short of the minimum length between the grips of the test apparatus (208\,mm), each end of a specimen is attached to 9-mm-wide glass-fiber reinforced polymer (GFRP) struts. Consequently, the protruding membrane width is cut off to minimize the distance from the X-ray system achieving high resolution of the computed tomography (CT) scan. The mechanical performance of the magnetic membranes was determined by conducting a combined experimental procedure of tensile testing with CT. The internal morphology and structure of the examined membranes was imaged in different levels of deformation with in situ CT scans. For this work, the CT-system FCTS 160 - IS by FineTec FineFocus Technologies was used. FCTS160-IS consists of the X-ray source FORE 160.01C TT and a flat panel-detector (2300\,px\,$\times$\,3200\,px), which both rotate around the specimen clamped on a ZwickRoell Z250 instrument. Displacement-controlled tensile tests with a nominal strain rate of 2.5\,mm/min were performed in combination with CT scans to determine the failure displacement of each membrane sample. The list of samples is provided in Supplementary Table~3.

\subsection{Dynamics of magnetic soft actuators in a magnetic field}
Electromagnetic coils were used to create an alternating magnetic field perpendicular to the magnetization direction of the sample. These alternating magnetic fields produced an oscillatory actuation of magnetic membranes. Depending on the intensity and frequency of the driving current supplying the coil, magnetic fields in the range from 0.5 to 5\,mT and frequencies from 1 to 20\,Hz were typically applied (unless is specified differently in specific cases along the text). Upon the change of polarity of the magnetic field in time, magnetic membranes reposition themselves making oscillatory movements. Each of the arms of the bow tie actuators were labeled with fluorescent inks for tracking purposes. Tracking the coordinates of the fluorescent marker in time allows to check the periodicity and harmonics of the signals {(Supplementary Movies S1 and S2)}. The Poincar\'{e} and spectral Fourier analysis allowed for a precise analysis of the obtained tracking signals to classify them as (i) periodic dynamics corresponding to actuation happening in correspondence with the frequency of the driving field; (ii) quasiperiodic with subharmonics, where actuation occurs repetitively with different frequency with respect to the main frequency of the driving field; and (iii) chaotic where the actuation pattern does not show periodicity. The reported video sequences were taken using a high-speed camera FASTCAM SA3 Model 120K from Photron, which can capture video with resolution of 1024\,px\,$\times$\,1024\,px at frame rates up to 2000 frames per second (fps) and at reduced resolution up to 120000\,fps but the memory of the camera is limited to 2\,GB. To perform long sequence continuous tracking for 40\,000 cycles, the soft actuators was operated for 35 min under a AC magnetic field with 6 mT amplitude and 19.5 Hz frequency. For this long sequence measurement, the video was recorded by a smartphone (Redmi K40, China) in slow motion mode (240 fps at resolution of 1280\,px $\times$ 720\,px).
%\highlight{To perform continuous tracking for more than 1 million points, the soft actuator was operated for about 6\,days at 20\,Hz. For these measurements, a video recording was not done due to constraints on the amount of generated data. Instead we used a Sensific's ODIN add-on solution for commercial microscopes. In addition to its regular video recording mode (210 fps at resolution of 1280\,px\,$\times$\,1024\,px and 10000\,fps at 400\,px\,$\times$\,32\,px), the camera has a real-time tracking function, which can track objects (e.g., fluorescent markers as in our case) at the rate of up to 10000\,fps. The tracked data was saved directly to a text file without recording or saving a video. In this way, the measurements with tracking 1 million cycles were recorded over a long period of time at a frame rate of 240\,fps, resulting in several gigabytes of tracked data.}
{For the dynamics experiments, the soft actuators are fixed in their middle point with a double sided tape. To avoid sticking of arms of the actuator to the supporting surface upon their actuation, it is important to appropriately choose the material of the substrate to assure its week adhesion to PDMS. In our case, we used regular paper as a support for our magnetic actuators. The actuation strategy for these ultrathin magnetic soft actuators is validated elsewhere~\cite{wang2020untethered}.}

\subsection{Stochastic computation demonstration}
We drove the {bow-tie-shape} soft robot in the chaotic regime at 35\,Hz and tracked the position of the fluorescent marker in the $x$ direction during 480\,s. The position of the marker was determined in units of pixels with a blur threshold equal to 50 using Open CV library~\cite{bradski2000opencv}. Frames without a well-established marker position were omitted from the analysis. The resulting array of coordinates is interpolated and sampled by the actuating field frequency to obtain Poincar\'{e} maps. For further processing, we used components of the Poincar\'{e} map. These arrays are stacked together for all recorded videos. We removed sequences of the sequential constant values originating from the cross-clapping motion and split the resulting array into sequences by 1500 numbers. For each of these sequences, 10~largest and 10~smallest numbers are removed and the rest of values are mapped to the range $[0,1]$ to obtain the dataset $X_i$ with $i$ being the set number. For $X_i$, the cumulative distribution function $\mathrm{CDF}_X$ is calculated, which reveals normal-like distribution of the initially obtained numbers. To obtain uniformly distributed random values, the dataset $Y_i=\mathrm{CDF}_X (X_i)$ is calculated. A comparison of the initial distribution of the generated numbers with the uniform distribution is shown in Supplementary Figure~14 using the so-called Q-Q (quantile-quantile) plot. A deviation of the quantiles from the diagonal line (shown in red) indicates that $X_i$ has a different to a uniform distribution. After transformation, $Y_i$ becomes the uniform distribution as it is shown in the inset of Figure~\ref{fig:stochastic_computing}c, where sample quantiles coincide with theoretical quantiles of the uniform distribution following the diagonal. Finally, all datasets are stacked together for further usage. We take each 5-th point from the dataset to obtain a sharp decay of the autocorrelation function shown in Figure~\ref{fig:stochastic_computing}d. To deal with $k$-bit non-negative integer numbers, the final dataset is scaled to the range $[0,2^{k-1}]$ with a floor function applied to the result.

The data was evaluated using a statistical test suite for random number generators from NIST~\cite{bassham2010statistical}. The random sequences generated by soft actuators driven in a chaotic regime successfully passed 14 tests. The random sequences are used as inputs for multiplying $42\times 54$ using a probabilistic algorithm schematized in Figure~\ref{fig:stochastic_computing}f. We implement the simplest version of the stochastic multiplication relying on the multiplication of probabilities. First, each decimal number is converted to a stochastic number by the following procedure~\cite{gross2019stochastic}. We consider only those numbers lying in range $[0,127]$ (7-bit numbers). Thus, each number $a\in [0,127]$ can be associated with the probability $p_a=a/127$. The $M$-bit stochastic number $S_M (a)$ associated with $a$ is represented by a sequence of $M$ zeros and ones, where the probability to pick ``1'' from the random position is equal to $p_a$. Having a sequence of random numbers $R_7$, $S_M (a)$ can be generated by making $M$ iterations over the sequence. At $i$-th iteration, $a$ is compared with $r_i$: if $a<r_i$ ($a>r_i$), then ``1'' (``0'') is used for $i$-th position in $S_M (a)$. The multiplication of two stochastic numbers $S_M (a)$ and $S_M (b)$ is performed as a logical AND operation over the respective binary sequences, which corresponds to the multiplication of probabilities $p_c=p_a\times p_b\equiv ab/127^2$, where $p_c$ is determined as the ratio between the number of ``1''s in the resulting sequence with respect to $M$. A conversion of the result back to the regular decimal number reads $c=127^2 p_c$.

\subsection{Reservoir computing demonstration}
The transformation task was performed using an input dataset consisting of 1000 data points and a lag value of 35. The MG time-series in the chaotic regime was generated by solving the equation \mbox{$\dfrac{dx(t)}{dt} =  \beta \dfrac{x(t-\tau)}{1 + \left[ x(t-\tau) \right]^{n}} - \gamma x(t)$} with $\beta$ = 0.2, $\gamma$ = 0.1, $\tau$ = 17, and $n$ = 10. The dataset consisted of 3000 data points was split into 80\% for training and 20\% for testing.
The MG time-series was transmitted at a rate of 960 data points per second. To synchronize the magnetic field input with the robot's motion, an LED with continuously varying brightness was incorporated in the setup. The position of a fluorescent marker was tracked using OpenCV~\cite{bradski2000opencv}, which extracted the $x(t)$ and $y(t)$ coordinates of the marker. These trajectories were normalized to the range $[0,1]$, prior to applying ridge regression. The ridge regression model used a regularization parameter of $\lambda = 0.0001$. All computations were carried out using the scikit-learn library~\cite{pedregosa2011scikit}.

\section*{Data availability}
All of the data supporting the conclusions are available within the article and the Supplementary Information. Additional data are available from the corresponding authors upon request. 

\bibliography{bibliography}

\section*{Acknowledgments}
Authors thank Ivan Antonenko and Prof. Jin Ge (HZDR) for their contribution at the initial stages of the project. This work was financed in part by the German Research Foundation (DFG) grants MA 5144/28-1, the ERC grant 3DmultiFerro (Project number:\,101141331) and European Commission (project REGO; ID:\,101070066). E. R., F.G. and G.F. acknowledge support from the project PRIN 20225YF2S4 – Magneto-Mechanical Accelerometers, Gyroscopes and Computing based on nanoscale magnetic tunnel junctions (MMAGYC) and the project PRIN 2020LWPKH7 - The Italian factory of micromagnetic modeling and spintronics, PON Capitale Umano (CIR\_00030) funded by the Italian Ministry of University and Research (MUR), and the PETASPIN association (www.petaspin.com).

\section*{Author contributions} 

E.S.O.-M., G.S.C.B., M.N.L. and X.W. fabricated samples. E.S.O.-M., L.G., M.N.L. and X.W. carried out experiments on soft robot dynamics. 
R.I., Y.Z. and L.G. developed videotracking setup.
G.T. and A.F. carried out computer tomography. 
G.T., A.F., G.S.C.B., M.N.L., F.G. and X.W. performed mechanical sample characterization.  
E.S.O.-M. and X.W. carried out magnetic sample characterization.
O.V.P. and G.M. performed video tracking and data analysis. 
O.V.P. and G.F. carried out data analysis and stochastic computing. 
E.R. and G.F. performed NIST tests and reservoir computing. 
E.S.O.-M. carried out eye blinking and digital voice modulation demonstrations. 
E.S.O.-M., O.V.P., G.F. and D.M. developed the concept. 
O.V.P., G.F. and D.M. lead the project. The manuscript was written by E.S.O.-M., O.V.P, E.R., D.M., G.F. with contribution of G.S.C.B., M.N.L., X.W., L.G., G.M., F.G., R.I., Y.Z., G.T, A.F.

\section*{Competing interests}
The authors declare no competing interests.

%%%%%%%%%%%%%%%% MAIN TEXT FIGURES %%%%%%%%%%%%%%%

\newpage

%\newpage
%\noindent

\newpage

\end{document}